\documentclass[letterpaper, 10 pt, conference]{ieeeconf}  

\IEEEoverridecommandlockouts                              

\usepackage{graphics} 
\usepackage{epsfig} 
\usepackage{mathptmx} 
\usepackage{times} 
\usepackage{amsmath} 
\usepackage{amssymb}  
\usepackage[T1]{fontenc}
\usepackage{booktabs} 
\usepackage{subcaption}
\usepackage{booktabs}
\usepackage{tabularx}
\usepackage{array}
\usepackage{multirow}

\title{\LARGE \bf
Constraint-Aware Diffusion Priors for High-Fidelity and Versatile Quadruped Locomotion}

\author{Jianhui Chen$^{\dagger}$, Ruixin Zhan$^{\dagger}$, Liu Liu$^{\ddagger}$, Yang Cai and Ziqiao Li
\thanks{This work was done in Amap, Alibaba Group.}
\thanks{$^{\dagger}$ These authors contributed equally to this work (Co-first authors).} 
\thanks{$^{\ddagger}$ Corresponding author.}
}

\begin{document}

\maketitle
\thispagestyle{empty}
\pagestyle{empty}

\begin{abstract}
Reinforcement learning combined with imitation learning has significantly advanced biomimetic quadrupedal locomotion. However, scaling these frameworks to massive, multi-source datasets exposes fundamental bottlenecks. First, traditional GAN-based discriminators are prone to mode collapse, struggling to capture diverse motion distributions from uncurated datasets. Second, existing kinematic priors suffer from out-of-distribution (OOD) tracking conflicts, leading to severe unintended heading drifts during complex maneuvers. Furthermore, deploying unconstrained priors to physical hardware poses critical safety risks by disregarding actuator dynamics. To overcome these challenges, we propose \textbf{Diff-CAST} (Diffusion-guided Constraint-Aware Symmetric Tracking), a novel motion prior framework leveraging the multi-modal distribution modeling capabilities of diffusion models for stylistic rewards. Diff-CAST effectively replaces traditional GAN discriminators, unlocking robust data scaling on heterogeneous collections. To ensure high-fidelity intent execution and reliable real-world deployment, we introduce a comprehensive Sim2Real architecture integrating Symmetric Augmented Command Conditioning (SACC) for drift-free tracking, and Constrained RL for hardware safety. Experiments on a quadruped demonstrate that Diff-CAST mitigates mode collapse, enables seamless transitions between diverse skills, and ensures robust, hardware-compliant locomotion.

\end{abstract}

\IEEEpeerreviewmaketitle

\section{Introduction}

Deep Reinforcement Learning (DRL) combined with Imitation Learning has significantly advanced biomimetic quadrupedal locomotion \cite{RoboImitationPeng20, hoeller2024anymal}. Within this domain, data-driven motion priors have emerged as the dominant paradigm. In particular, Adversarial Motion Priors (AMP) \cite{peng2021amp, escontrela2022adversarial} employ a GAN discriminator to extract stylistic rewards from unannotated motion capture (mocap) datasets, synthesizing natural gaits without exhaustive handcrafted reward engineering. Recently, diffusion models have also shown promise in the capture of complex multi-skill locomotion datasets \cite{huang2024diffuseloco}.

However, scaling standard adversarial frameworks to heterogeneous, multi-source datasets exposes three fundamental bottlenecks:
\begin{enumerate}
\item \textbf{Limited Motion Diversity:} Traditional GAN discriminators frequently exploit gradient shortcuts, over-optimizing for dominant modes (e.g., forward running) while forgetting minority skills \cite{margolis2022walktheseways}. Recent multi-skill extensions \cite{vollenweider2023advanced, huang2025learning} rely on laborious manual phase annotations, struggling to model fluid interpolations for seamless gait transitions.
\item \textbf{Inaccurate Command Tracking:} In OOD scenarios (e.g., lateral stepping), unconditioned priors heavily penalize novel kinematics, overriding directional commands to pull behaviors back toward the forward-biased training manifold \cite{kumar2020conservative}. This manifests physically as persistent heading drifts and unintended trajectory curvatures.
\item \textbf{Unsafe Sim-to-Real Deployment:} Purely kinematic-driven priors neglect actuator dynamics. Unconstrained policies often generate aggressive control signals, triggering dangerous torque spikes and joint velocity saturation upon physical deployment \cite{yang2022safe, chane2024soloparkour}.
\end{enumerate}

\begin{figure}[t]
    \centering
    \includegraphics[width=\linewidth]{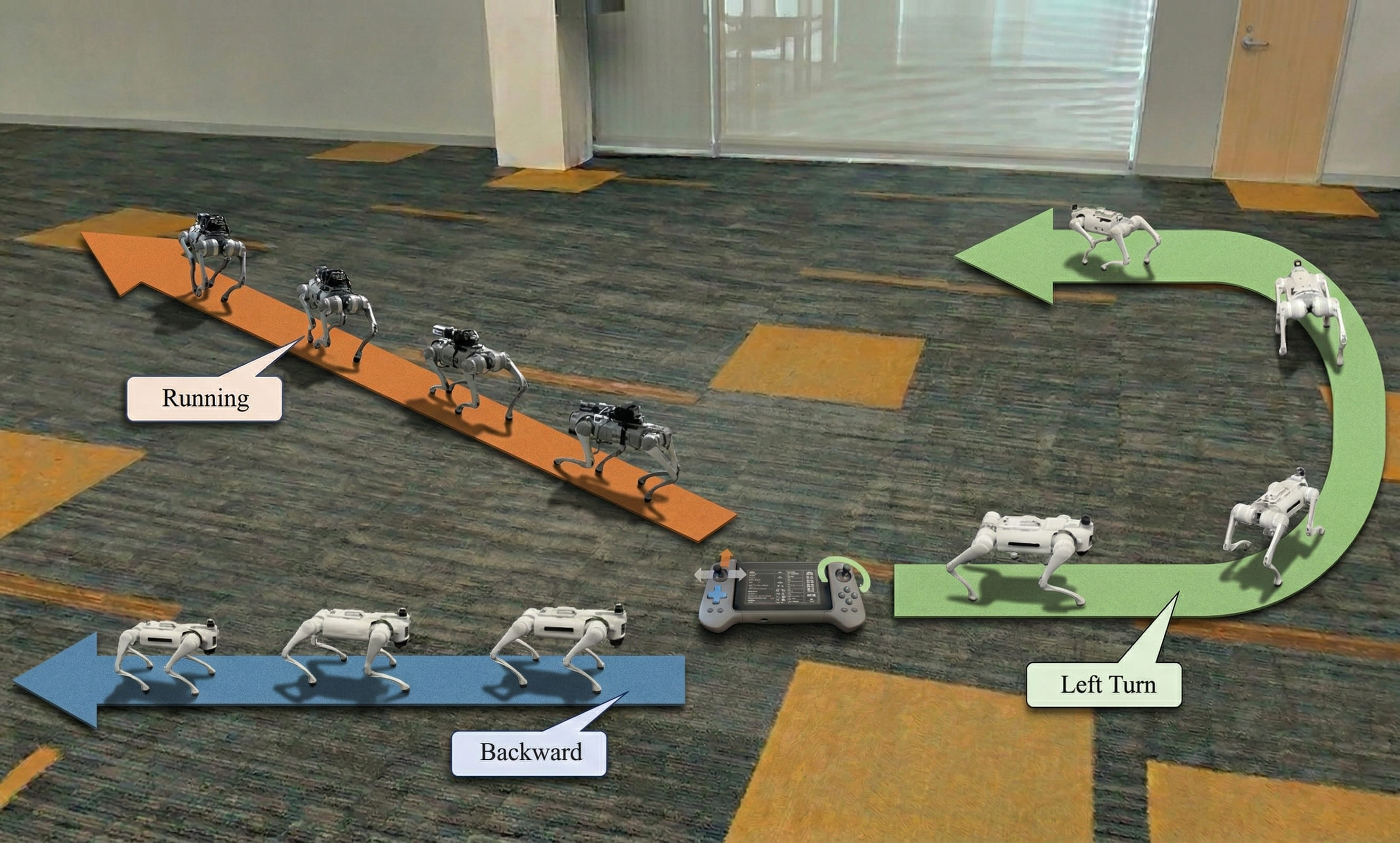}
    \caption{\textbf{Real-world deployment of Diff-CAST.} Our constraint-aware diffusion prior allows the physical robot to safely perform complex omnidirectional maneuvers and seamlessly transition across diverse gaits without heading drift.}
    \label{fig:teaser}
\end{figure}

To surmount these challenges, we propose \textbf{Diff-CAST}, a diffusion-guided motion prior framework with a robust sim-to-real architecture. Instead of a binary classifier, Diff-CAST leverages the conditional denoising reconstruction error of diffusion models to formulate continuous stylistic rewards, naturally accommodating heterogeneous unlabelled datasets and mitigating mode collapse. We resolve OOD tracking conflicts via a decoupled command-conditioned balancing mechanism. Finally, we ensure reliable physical deployment by enforcing hardware safety through Constrained RL and a transient-aware command curriculum.
The primary contributions of this paper are summarized as follows:
\begin{itemize}
    \item We propose \textbf{Diff-CAST}, a diffusion-guided motion prior method enhanced with an action-agnostic formulation and a bounded reward. This approach substantially mitigates mode collapse and enables zero-shot, seamless multi-modal skill transitions within a single policy.
	\item We propose \textbf{Symmetric Augmented Command Conditioning} to mathematically resolve the task-style objective mismatch in OOD scenarios. SACC guarantees precise omnidirectional command tracking and reduces real-world heading drifts prevalent in baseline methods to a negligible level.
    \item We propose a comprehensive Safe Sim-to-Real architecture tailored for generative motion priors. By integrating Constrained RL with a transient-aware command curriculum, our framework achieves hardware-safe, dynamically balanced, and versatile locomotion on a physical quadruped.
\end{itemize}

	\section{Related work}
	
	\subsection{Imitation Learning for Locomotion}
	Synthesizing highly natural biomimetic locomotion has long been driven by imitation learning combined with DRL. Early pioneering frameworks, such as DeepMimic \cite{peng2018deepmimic}, achieved impressive results through explicit kinematic tracking, but typically learned isolated skills with limited environmental robustness. To overcome the constraints of rigid tracking, AMP introduced a paradigm shift utilizing GAN discriminators to provide flexible stylistic rewards \cite{peng2021amp}. This adversarial approach has been widely deployed in physical quadrupeds \cite{escontrela2022adversarial} and has expanded to latent space architectures for complex skill discovery, such as ASE \cite{peng2022ase} and CALM~\cite{tessler2023calm}.
	
    However, scaling standard MLP-based adversarial priors exposes critical limitations. They are highly susceptible to mode collapse \cite{li2023learning}, over-optimizing for dominant behaviors. Extensions like Multi-AMP \cite{vollenweider2023advanced} and CAMP \cite{huang2025learning} require manual data segmentation, struggling with continuous zero-shot transitions on uncurated datasets. Furthermore, unconditioned priors heavily penalize novel kinematics during OOD maneuvers, overriding commands and inducing persistent heading drifts \cite{krueger2021out}. Without explicit symmetry regularizations, policies overfit to dataset biases, causing asymmetric gaits \cite{mittal2024symmetry, nie2025coordinated}. Physically, purely kinematic frameworks ignore actuator dynamics \cite{hwangbo2019learning}, generating aggressive signals that cause dangerous torque spikes upon real-world deployment \cite{chane2024soloparkour, yang2022safe, as2025spidr}.

    In contrast, Diff-CAST leverages the continuous, label-free gradient field of diffusion models \cite{huang2024diffuseloco} to natively encapsulate multi-modal manifolds. By integrating this prior directly into the high-frequency RL loop alongside a robust sim-to-real safety bundle, Diff-CAST overcomes mode collapse, heading drifts, and hardware limits.

\subsection{Diffusion Models in RL and Robotics}
Denoising Diffusion Probabilistic Models (DDPMs) exhibit exceptional multi-modal modeling capabilities. Generative policies \cite{janner2022planning, chi2025diffusion} and RL-steered diffusion frameworks \cite{wagenmaker2025steering, hansen2023idql} reframe decision-making as a conditional generative process. However, the iterative Langevin dynamics required for action sampling incur severe inference latency, rendering them incompatible with the high-frequency (e.g., 50Hz) control loops mandated for dynamic quadrupedal balancing.

Alternatively, recent works leverage diffusion models strictly as dense reward signals \cite{lai2024diffusion, huang2024diffusion, urain2022se3dif}. Yet, methods like DRAIL \cite{lai2024diffusion} rely on state-action pairs $(s,a)$—preventing the use of action-free biological mocap data—and utilize unbounded rewards ($r = \log(D) - \log(1-D)$) that trigger value network explosions during Proximal Policy Optimization (PPO) updates. Diff-CAST advances this paradigm through an action-agnostic state-transition formulation and a bounded reward manifold. Crucially, we utilize the diffusion model exclusively during offline training to shape the stylistic energy landscape. This decoupling retains deep distribution-matching capabilities while operating online as an ultra-fast MLP, strictly preserving the zero-latency execution mandated by physical quadrupeds.

\begin{figure*}[t]
	\centering
 	\includegraphics[width=1\linewidth]{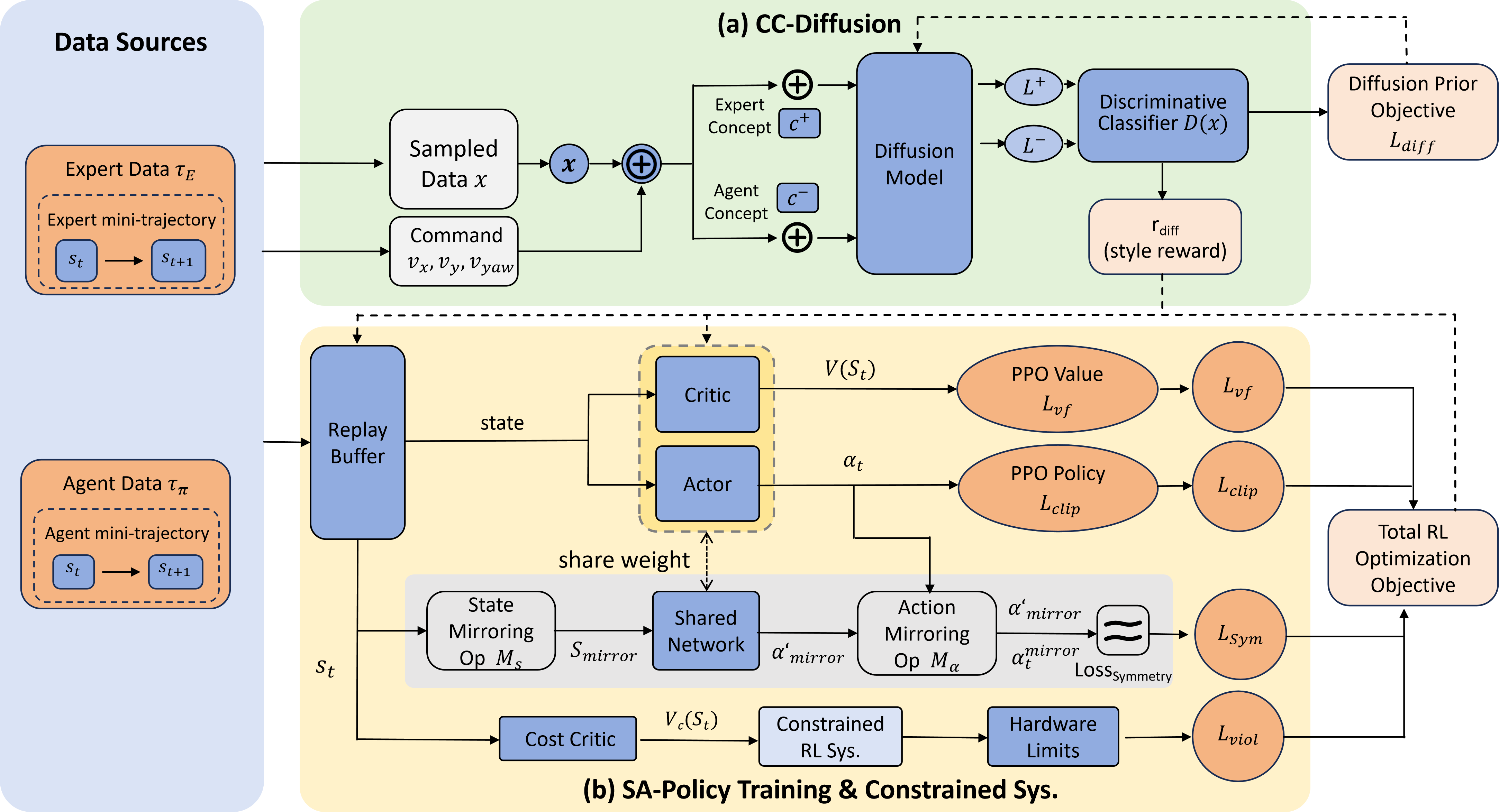}
    \caption{\textbf{Overview of the Diff-CAST framework.} The method consists of \textbf{(a) CC-Diffusion}, which learns a command-conditioned diffusion prior from expert and agent transitions and derives a stylistic reward for policy learning, and \textbf{(b) SA-Policy Training \& Constrained System}, which integrates PPO optimization, symmetry-aware regularization, and constrained RL to achieve stable, balanced, and hardware-safe locomotion.}
    \label{fig:system-overview}
    \end{figure*}

\section{METHODOLOGY}

The primary objective of our framework is to learn a unified control policy capable of seamlessly executing a diverse repertoire of biomimetic motor skills (e.g., walking, running, and lateral stepping) while strictly adhering to the electromechanical limits of physical hardware.

\subsection{System Overview}
We formulate the quadrupedal locomotion control problem as a discrete-time Markov Decision Process (MDP) defined by the tuple $\langle S, A, P, R, \gamma \rangle$. At each control step $t$, a lightweight actor policy $\pi_\theta(a_t | s_t, v^{cmd})$ maps the current proprioceptive state $s_t \in S$ and a user-specified velocity command $v^{cmd} \in \mathbb{R}^3$ to target joint positions $a_t \in A$. To circumvent the need for exhaustive manual reward engineering, we decompose the total reward $r$ into two complementary components:
\begin{equation}
    r = \omega_{task} r_{task} + \omega_{diff} r_{diff},
\end{equation}
where $r_{task}$ encourages the precise tracking of velocity commands, and $r_{diff}$ is a data-driven stylistic reward evaluated by an enhanced diffusion-based adversarial prior.

As illustrated in Fig.~\ref{fig:system-overview}, our framework builds upon and significantly adapts the AMP paradigm for high-frequency legged control. The architecture is structurally decoupled into two primary workflows: 
\begin{itemize}
    \item \textbf{CC-Diffusion (Fig.~\ref{fig:system-overview}(a)):} Operating independently of the high-frequency control loop, this module learns a command-conditioned diffusion prior from expert and agent state transitions. By incorporating explicit velocity commands together with domain concepts, it extracts intent-aligned motion features and provides a stylistic reward that distinguishes expert-like behaviors from the agent's rollouts.
    \item \textbf{SA-Policy Training \& Constrained Sys. (Fig.~\ref{fig:system-overview}(b)):} During online reinforcement learning, the policy is optimized with PPO under the guidance of the frozen diffusion-based stylistic reward. In parallel, a symmetry-aware branch regularizes the policy to preserve morphological consistency, while a constrained RL branch enforces hardware-related safety limits for reliable sim-to-real deployment. Together, these components define the overall RL training objective.
\end{itemize}

The structurally decoupled architecture is motivated by the algorithmic and hardware limitations of naively deploying generative priors for physical legged locomotion. The following subsections detail the \textit{Action-Agnostic and Bounded Diffusion Prior} (Sec. III-B) to stabilize PPO, \textit{SACC} (Sec. III-C) to resolve OOD tracking conflicts, and our \textit{Safe Sim-to-Real} architecture (Sec. III-D) for physical deployment.

\subsection{Action-Agnostic and Bounded Diffusion Prior}

While emerging diffusion-based adversarial frameworks (e.g., DRAIL \cite{lai2024diffusion}) effectively capture multi-modal distributions, their naive deployment in physical locomotion exposes two critical bottlenecks: strict action dependency and value network instability caused by unbounded rewards. We resolve these via an action-agnostic state-transition classifier and a strictly bounded reward formulation.

\subsubsection{Action-Agnostic State-Transition Prior}
Conventional adversarial methods, including DRAIL, evaluate state-action pairs $(s_t, a_t)$. However, acquiring high-fidelity torque data from biological motion capture is impractical, and the inherent morphological sim-to-real gap renders action-conditioned imitation exceptionally brittle. 

To bypass this, we eliminate the action space entirely, shifting the generative modeling exclusively to temporal state transitions $x_t=(s_t,s_{t+1})$. As illustrated in Fig.~\ref{fig:system-overview}(a), this action-agnostic architecture acts as a continuous kinematic prior. This decouples learning from the restrictive actuator domain, enabling robust cross-embodiment retargeting.

\subsubsection{Bounded Diffusion Reward via Conditional Denoising}
Standard adversarial methods derive stylistic rewards using an unbounded log-likelihood ratio, $r=\log(D)-\log(1-D)$. Even recent diffusion variants rigidly apply this transformation. During early exploration, an agent inevitably deviates from the expert manifold ($D \to 0$), triggering extreme negative reward spikes ($r \to -\infty$). This severely destabilizes the PPO value network and precipitates policy collapse.

To guarantee stability, we utilize a conditional denoising network $\epsilon_\varphi(x_{t,k}, c, k)$ to output a bounded probability. Here, $k$ is the timestep drawn via antithetic sampling, $x_{t,k}$ is the noise-injected transition, and $c \in \{c^+, c^-\}$ represents the domain concept ($c^+$ for expert, $c^-$ for agent), perfectly aligning with Fig.~\ref{fig:system-overview}(a). We compute the denoising Mean Squared Error (MSE) under both hypotheses:
\begin{align}
    L^+(x_t) &= \mathbb{E}_{k,\epsilon} \left[ \|\epsilon - \epsilon_\varphi(x_{t,k}, c^+, k)\|^2 \right], \\
    L^-(x_t) &= \mathbb{E}_{k,\epsilon} \left[ \|\epsilon - \epsilon_\varphi(x_{t,k}, c^-, k)\|^2 \right].
\end{align}

Since the diffusion denoising error bounds the negative log-likelihood ($L \propto -\log p(x|c)$), we analytically derive the classification probability $D_\varphi(x_t) \in [0, 1]$ via a softmax function:
\begin{equation}
    D_\varphi(x_t) = \frac{\exp(-L^+(x_t))}{\exp(-L^+(x_t)) + \exp(-L^-(x_t))}.
\end{equation}

To optimize the diffusion parameters $\varphi$, we minimize the Binary Cross-Entropy (BCE) loss between the expert ($\tau_E$) and agent ($\tau_\pi$) buffers, yielding the objective $L_{\text{diff}}$ (Fig.~\ref{fig:system-overview}(a)):
\begin{equation}
    L_{\text{diff}}(\varphi) = \mathbb{E}_{\tau_E}[-\log D_\varphi(x)] + \mathbb{E}_{\tau_\pi}[-\log(1 - D_\varphi(x))].
\end{equation}

Crucially, bypassing the volatile unbounded logit transformation used in prior arts, we formulate our stylistic reward $r_{\text{diff}}$ directly as this bounded probability:
\begin{equation}
    r_{\text{diff}} = D_\varphi(x_t).
\end{equation}

This principled modification retains mode-seeking gradient guidance while structurally bounding the reward magnitude within $[0, 1]$, completely preventing extreme penalty spikes and ensuring robust policy convergence during hardware training.

\subsection{Omnidirectional Command Tracking via SACC}
Biological datasets are intrinsically forward-biased and asymmetric. Deploying unconditioned generative priors for omnidirectional control internalizes these flaws, causing the policy to prioritize stylistic rewards over user commands. This exposes a dual out-of-distribution (OOD) challenge:
\begin{itemize}
\item \textbf{Directional Bias:} When commanded to walk perfectly straight, an unconditioned prior penalizes this symmetric motion due to inherent dataset asymmetries. This drags the policy toward biased expert manifolds, resulting in persistent heading drifts.
\item \textbf{Kinematic Sparsity:} Datasets fundamentally lack pure lateral maneuvers. The prior heavily penalizes novel kinematics like lateral stepping, compelling the policy to override user commands and inducing unintended trajectory curvatures.
\end{itemize}

To resolve this conflict, SACC couples an intent-conditioned diffusion prior with morphology-aware geometric regularizations.

\begin{figure}[h]
    \centering
    \includegraphics[width=\linewidth]{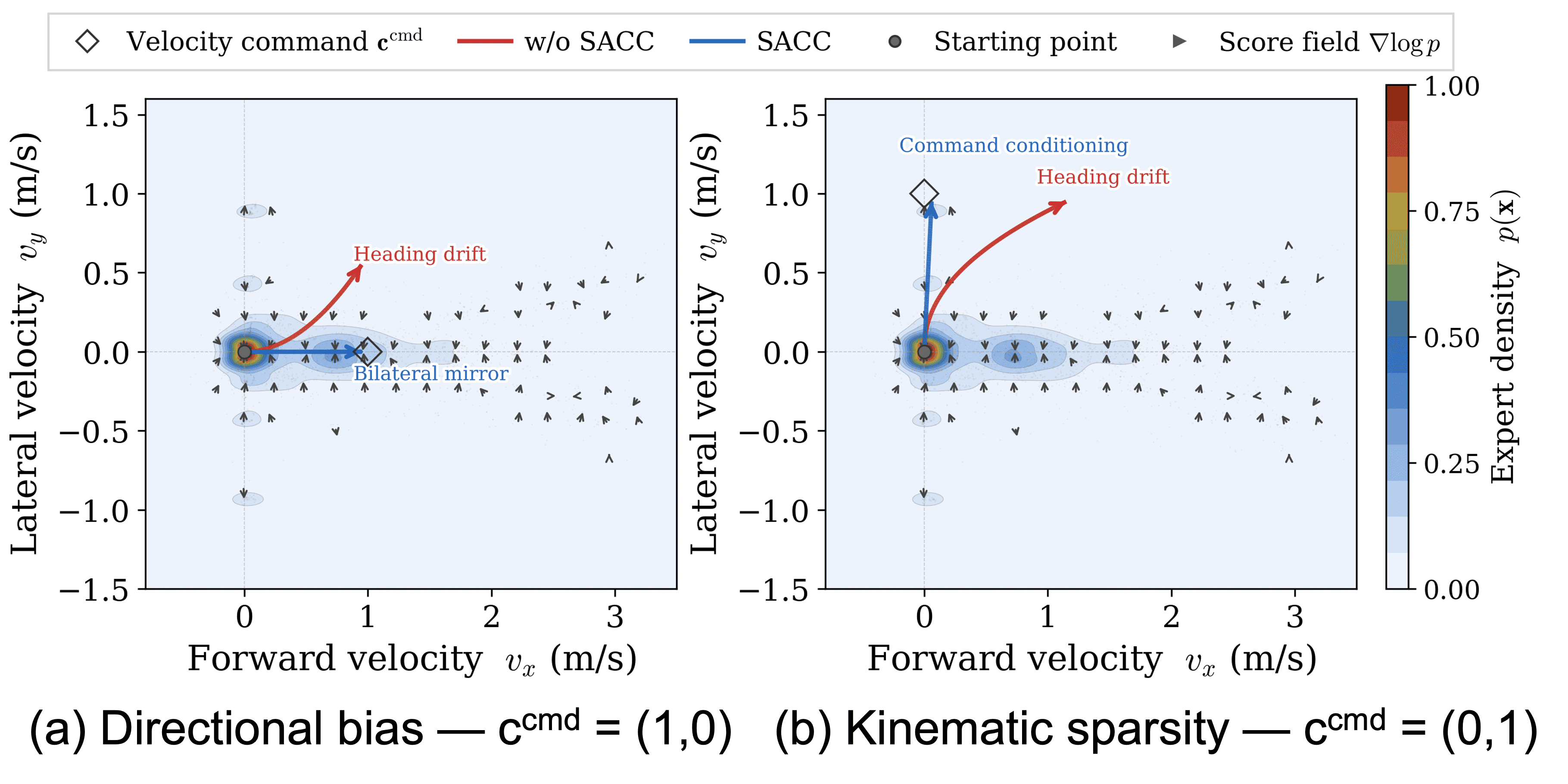}
\caption{OOD challenges in omnidirectional command tracking with unconditioned motion priors. The heatmap shows expert density $p(\mathbf{x})$, and arrows denote the score field $\nabla \log p$.}
    \label{fig:sacc}
\end{figure}

\subsubsection{Resolving Command Override via Intent Conditioning}
To address the sparsity-induced objective conflict (Challenge 2) and prevent the discriminator from penalizing novel kinematics against a marginal forward prior, we expand diffusion conditioning to explicitly include the real-time velocity command $v^{cmd} \in \mathbb{R}^3$. Because unannotated datasets lack explicit intent labels, we analytically extract localized torso velocities from the expert trajectories to serve as pseudo-commands $v^{cmd}_E$ via hindsight relabeling. All kinematic features are projected into a heading-decoupled local base frame.

To satisfy the strict latency constraints of closed-loop control, the continuous intent $v^{cmd}$ is directly concatenated with the noise-injected transition $x_{t,k}$ via an early-fusion architecture. This transforms the marginal diffusion prior into a conditional denoising network: $\epsilon_\varphi([x_{t,k}, v^{cmd}], c, k)$. Consequently, a lateral step is evaluated against the mathematically aligned lateral expert sub-manifold, effectively eliminating the command-override dilemma.

\subsubsection{Resolving Bias and Sparsity via Geometric Regularizations}
While command conditioning aligns the evaluation objective, purely data-driven Multi-Layer Perceptrons (MLPs) cannot autonomously synthesize lateral data that is fundamentally absent, nor can they eradicate inherent directional biases (Challenge 1). To physically ground the policy and synthetically expand the expert support, we enforce spatial invariance across two dimensions:

\begin{itemize}
\item \textbf{Kinematic Symmetry:} We define a sagittal-plane mirror operator $M(\cdot)$ that swaps contralateral legs and negates asymmetric spatial components (lateral velocity, roll and yaw rate). Augmenting the offline dataset with $M(\cdot)$ perfectly balances unilateral biases and synthesizes missing lateral counterparts. Concurrently, we impose a structural mirror symmetry loss on the actor-critic networks to prevent unilateral limping. Let $\tilde{s}_t = M_s(s_t)$ and $\tilde{v}^{cmd} = M_v(v^{cmd})$ denote the mirrored observation and velocity command, respectively:
\begin{equation}
	\label{eq:sym-losss}
	\begin{aligned}
		L_{sym} = \lambda \Big( 
		&\big\|\pi_\theta(s_t, v^{cmd}) - M_a\big( \pi_\theta(\tilde{s}_t, \tilde{v}^{cmd}) \big) \big\|^2 \\
		+\; &\big\|V_\psi(s_t, v^{cmd}) - V_\psi(\tilde{s}_t, \tilde{v}^{cmd}) \big\|^2 
		\Big),
	\end{aligned}
\end{equation}
where $M_a$ denotes the specific mirror operator for the action space. The target value $V_\psi$ is detached to stabilize optimization.
	
	\item \textbf{Yaw Invariance:} Quadrupedal locomotion is inherently independent of the absolute global heading. To prevent the discriminator from acting as a directionally biased prior and to artificially synthesize $360^\circ$ rotational data, heading-dependent planar features are dynamically rotated by a uniformly sampled yaw offset $\delta \sim U(-\pi, \pi)$ during the diffusion update.
\end{itemize}

Crucially, co-optimizing these geometry-aware mechanisms successfully anchors early exploration to valid, unbiased expert manifolds. This prevents gradient conflicts and ensures precise, omnidirectional locomotion without drift.

\subsection{Hardware-Safe Sim-to-Real Deployment}
For quadrupeds like the Unitree Go2, standard Domain Randomization (DR) typically suffices for normal scenarios \cite{kumar2021rma, zhang2025track, huang2025barlowwalk}. However, purely data-driven priors neglect electromechanical constraints. By greedily maximizing stylistic rewards, unconstrained policies trigger actuator limit violations and torque oscillations that DR cannot reliably mitigate. To ensure hardware safety without heuristically distorting the policy, we formulate the control problem as a Constrained Markov Decision Process (CMDP). We adopt the Normalized Penalized Proximal Policy Optimization (N-P3O) scheme\cite{lee2024exploring} to bound actions and stabilize training via normalized cost advantages. Accordingly, we enforce joint position, velocity, and torque limits through an adaptive asymmetric penalty:
\begin{equation}
	\label{eq:cost-viol}
	L_{\text{viol}} = \sum_{i=1}^3 \lambda_i^{(t)} \max \left( 0, \, C_{\text{surr}}^{(i)} + \bar{v}_i \right),
\end{equation}
where $C_{surr}^{(i)}$ denotes the surrogate objective of the $i$-th constraint cost evaluated using normalized cost advantages, $\bar{v}_i$ is the constraint margin term (which incorporates the allowable safety limits and advantage normalization statistics), and $\lambda_i^{(t)}$ is an exponentially increased penalty coefficient scheduled to promote early-stage exploration. Owing to the ReLU gating, the penalty contributes no gradient when $C_{surr}^{(i)} + \bar{v}_i \le 0$, thereby preserving optimization capacity for the primary biomimetic objective in the feasible regime.

To reduce physically implausible transient behaviors and ensure stability under abrupt command changes (e.g., emergency stops), we introduce a transient-aware command curriculum. Specifically, discontinuous velocity commands are smoothed via a first-order Exponential Moving Average (EMA) to respect actuator bandwidth limits. Additionally, target commands are overridden to zero near the end of each training episode. This approach inherently encourages the policy to learn compliant deceleration and maintain a stable resting posture, thereby ensuring safer physical deployment.

\section{EXPERIMENTS}

To evaluate the proposed framework, our experiments are designed to answer the following three primary research questions (RQ).
\begin{itemize}
    \item \textbf{RQ1 (Motion Diversity):} Can diffusion prior effectively overcome the mode collapse problem inherent in traditional adversarial approaches (e.g., AMP) and synthesize diverse, multi-modal locomotion behaviors?
    \item \textbf{RQ2 (Command Tracking Accuracy):} How does the proposed \textit{Symmetric Augmented Command Conditioning} resolve the objective conflict in OOD kinematic states, thus preventing heading drifts and ensuring highly precise omnidirectional command tracking?
    \item \textbf{RQ3 (Sim-to-Real Safety):} Does the integration of Constrained RL and transient-aware command curriculum successfully bridge the sim-to-real gap, ensuring training stability and hardware-safe deployments, particularly during highly dynamic motions?
\end{itemize}

\subsection{Experimental Setup}
We train our policies in NVIDIA Isaac Gym using the Unitree Go2 quadruped robot. The expert dataset comprises $\sim\! 10$ minutes of completely label-free MoCap clips, encompassing a diverse range of behaviors such as forward locomotion, random exploratory movements, and turning maneuvers across mixed gaits. Notably, this raw dataset is inherently forward-biased, lacking pure backward maneuvers. 

Diff-CAST (\textbf{Ours}) is compared against a primary baseline, \textbf{Vanilla AMP} (using an MLP discriminator without diffusion), alongside two ablated variants: 1) \textbf{w/o SACC} (removing random rotation augmentations and symmetry constraints); and 2) \textbf{w/o Safe Sim-to-Real} (removing Constrained RL and the transient-aware command curriculum, only relying on standard Domain Randomization). We select Vanilla AMP as our primary baseline because it strictly shares our uncurated, zero-annotation problem setting. Conversely, recent multi-skill extensions (e.g., \cite{vollenweider2023advanced, huang2025learning}) fundamentally rely on manual phase annotations or dataset pre-segmentation, making them incompatible with our fully label-free training pipeline.

\begin{figure}[h]
    \centering
    
    \begin{subfigure}{\linewidth}
        \centering
        \includegraphics[width=\linewidth]{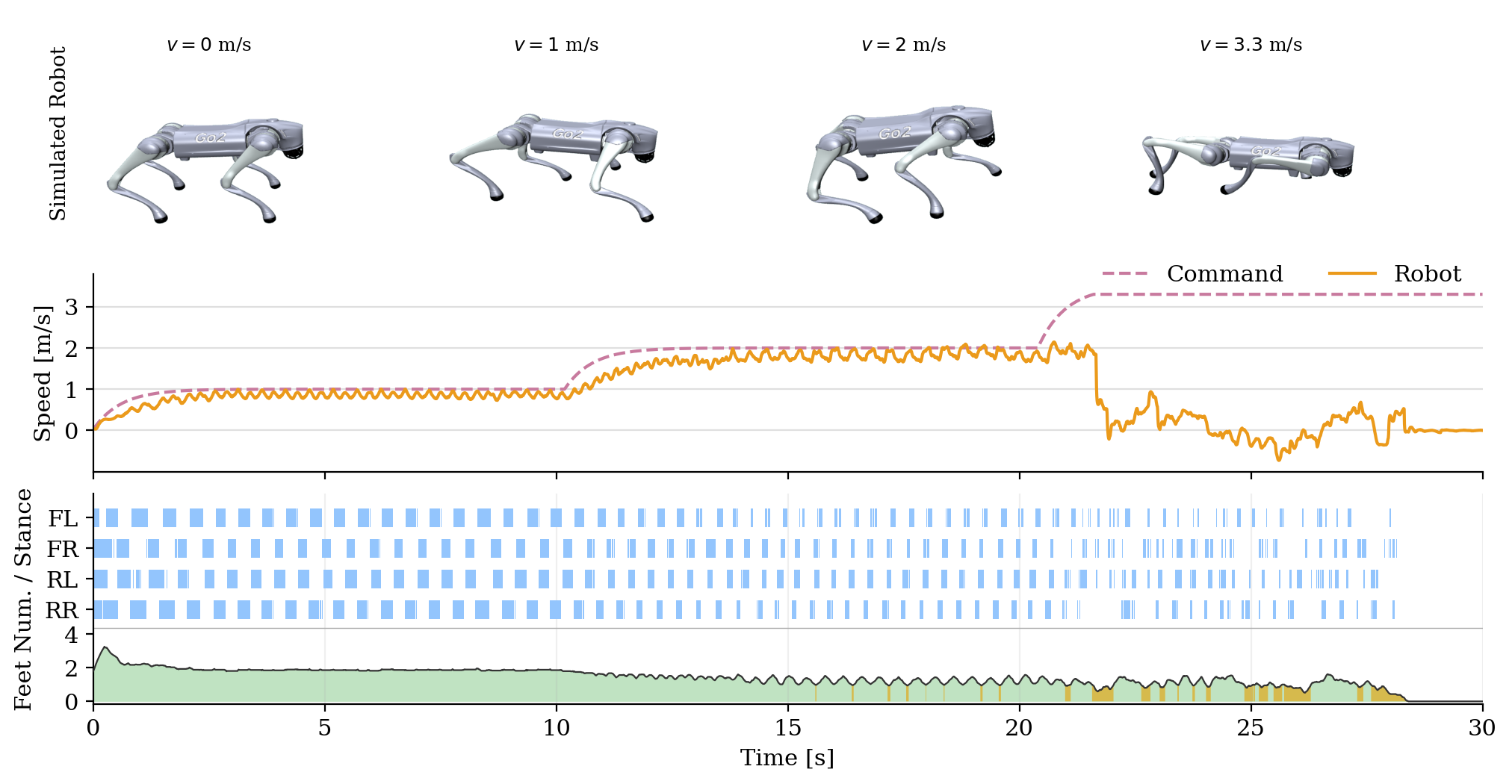}
        \caption{Vanilla AMP}
        \label{fig:ampold_pose}
    \end{subfigure}
    \begin{subfigure}{\linewidth}
        \centering
        \includegraphics[width=\linewidth]{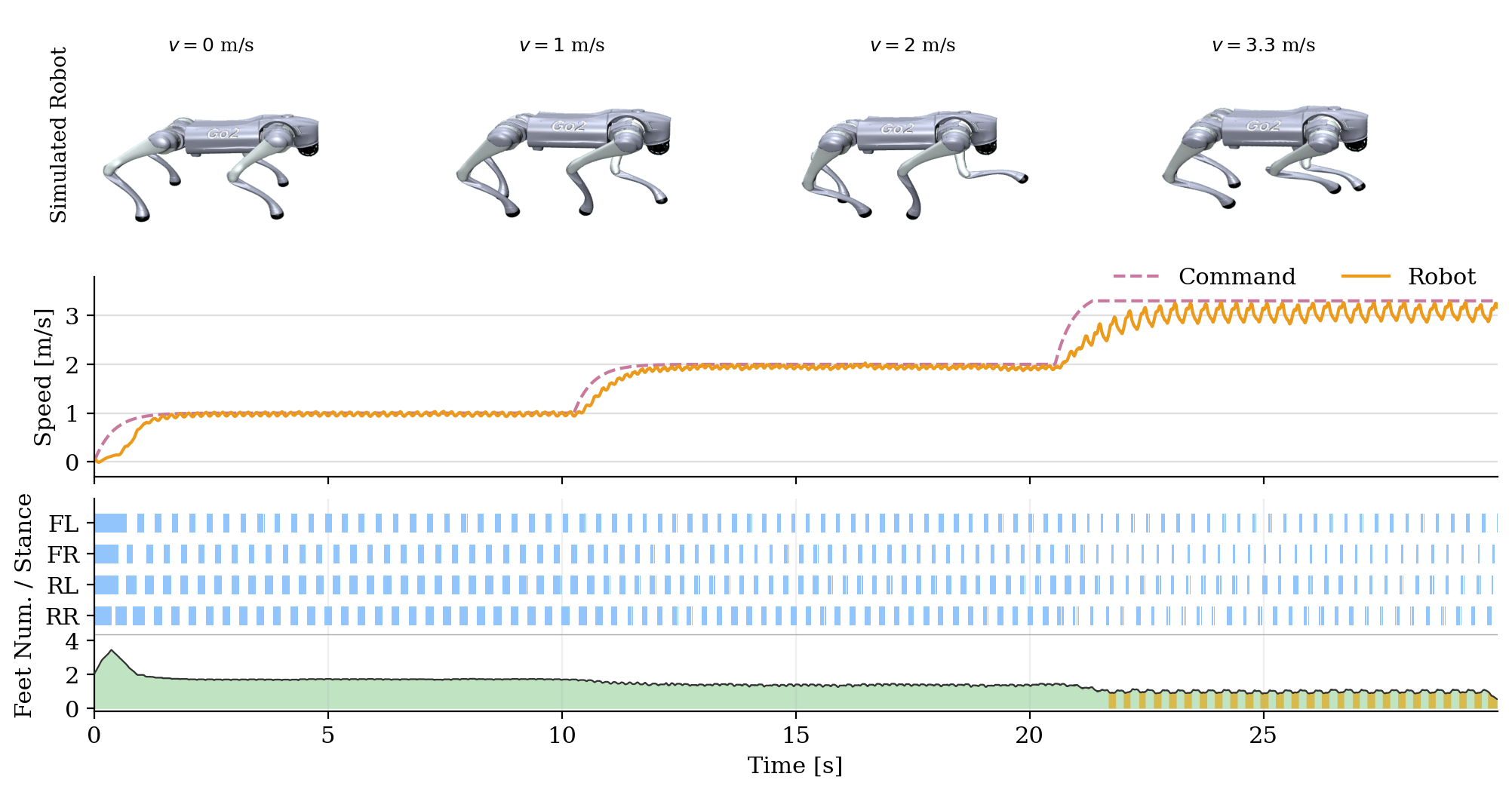}
        \caption{Diff-CAST}
        \label{fig:amp_pose}
    \end{subfigure}

\caption{\textbf{Footfall phases with support statistics.} Stance phases are shown per leg (blue), and the bottom bar summarizes the instantaneous number of supporting feet. (a) Vanilla AMP breaks down under high-speed commands, exhibiting disrupted contact patterns and reduced support. (b) Diff-CAST transitions gaits conditioned on the commanded velocity and maintains stable ground contact across speeds.}
    \label{fig:pace_transition}
\end{figure}

\subsection{Evaluation of Motion Diversity and Gait Realism}
Traditional adversarial imitation learning equipped with conventional MLP discriminators frequently suffers from mode collapse when exposed to massive, un-segmented, multi-modal MoCap datasets. In this section, we evaluate how our continuous diffusion prior fundamentally resolves this bottleneck to unlock scalable motion diversity, answering RQ1.


\textbf{Fluid Skill Transitions.} Mode collapse physically manifests as rigid gait timing. As illustrated in Fig.~\ref{fig:pace_transition}(a), Vanilla AMP fails to differentiate locomotion modes, maintaining an invariant ground-contact rhythm regardless of the commanded speed. In contrast, in Fig.~\ref{fig:pace_transition}(b), Diff-CAST seamlessly transitions across distinct, biomechanically accurate gait patterns conditioned on real-time target velocities: a four-beat walk at low speeds ($<1.0$ m/s), a diagonal trot at medium speeds ($\sim 2.0$ m/s), and a synchronized bound at high speeds ($>3.0$ m/s). Crucially, the continuous score-matching energy landscape of the diffusion prior enables fluid kinematic interpolations, allowing the policy to execute spontaneous, seamless transitions between these diverse skills without requiring heuristic phase oscillators.

\begin{figure}[h]
    \centering
    \includegraphics[width=\linewidth]{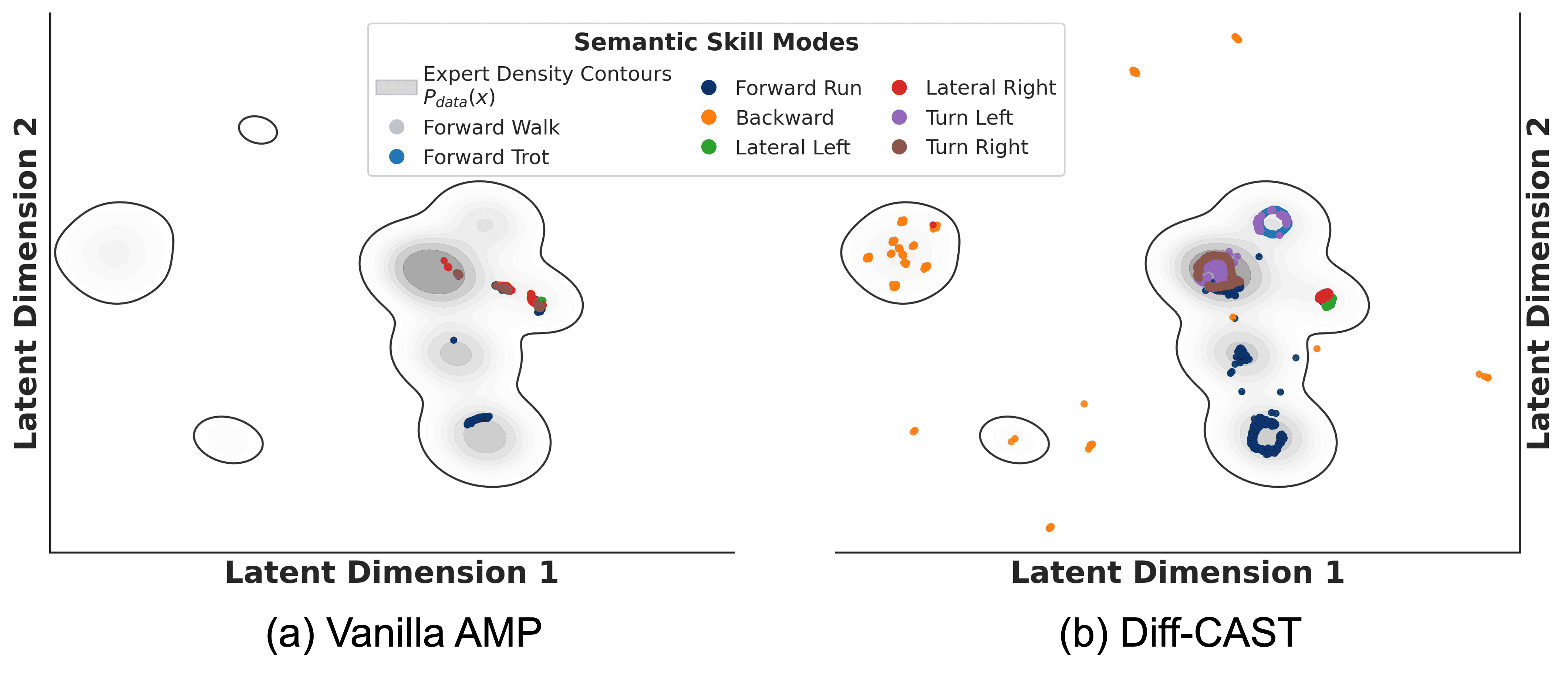}
    \caption{\textbf{Latent dimension analysis via UMAP.} Vanilla AMP (a) exhibits severe mode collapse, whereas Diff-CAST (b) successfully disentangles diverse semantic skills and autonomously synthesizes novel backward maneuvers.}
    \label{fig:umap_manifold}
\end{figure}

\textbf{Latent Space Analysis and Quantitative Realism.} To further investigate this multi-modal disentanglement, we project the learned behavior embeddings into a 2D latent space via UMAP \cite{healy2024uniform}. As shown in Fig.~\ref{fig:umap_manifold}(a), Vanilla AMP exhibits severe mode collapse, blending distinct gaits into an indistinguishable, entangled cluster. In contrast, our Diff-CAST (Fig.~\ref{fig:umap_manifold}(b)) acts as a high-fidelity manifold regularizer, successfully separating the diverse semantic skills into well-defined, localized regions. Quantitatively, we measure motion realism using the Fr\'echet Gait Distance (FGD) \cite{yoon2020speech, tevet2023human}, which evaluates the distributional divergence between the expert and generated kinematics. Our method achieves an FGD of 489.13, outperforming Vanilla AMP (4173.66). (A summary of quantitative metrics is deferred to the ablation study in Section \ref{subsec:quantitative_ablations}). This quantitative improvement confirms that the conditional denoising mechanism effectively preserves the rich intra-mode variations of the expert manifold.

\textbf{Data Scalability and Zero-Shot Skill Emergence. } Unlike traditional adversarial methods requiring curated data, our Diff-CAST robustly scales to massive, unlabelled MoCap datasets. Furthermore, it enables zero-shot emergence of novel skills by autonomously synthesizing backward maneuvers entirely absent from the expert data. As visualized in Fig. 5(b), these generated backward motions manifest as orange outliers outside the expert density contours, explicitly demonstrating the model's capacity to extrapolate beyond the training distribution.

\subsection{Evaluation of Command Tracking}
\begin{figure}[h!]
    \centering
    \begin{subfigure}[h]{0.5\linewidth}
        \centering
        \includegraphics[width=\linewidth]{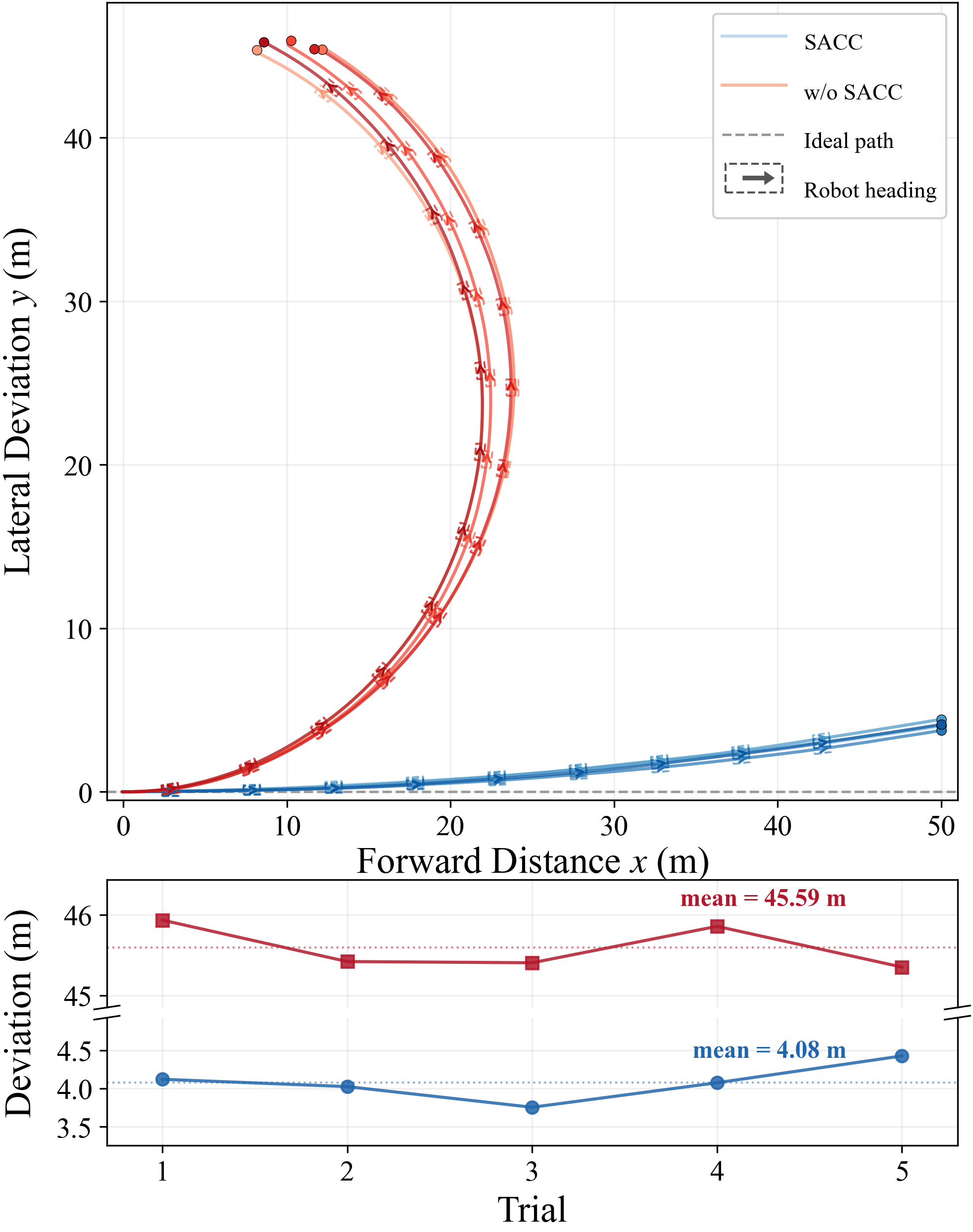}
        \caption{Straight-line tracking error along the $x$ direction (1m/s).}
        \label{fig:traj_heading_x}
    \end{subfigure}
    \hfill
    \begin{subfigure}[h]{0.47\linewidth}
        \centering
        \includegraphics[width=\linewidth]{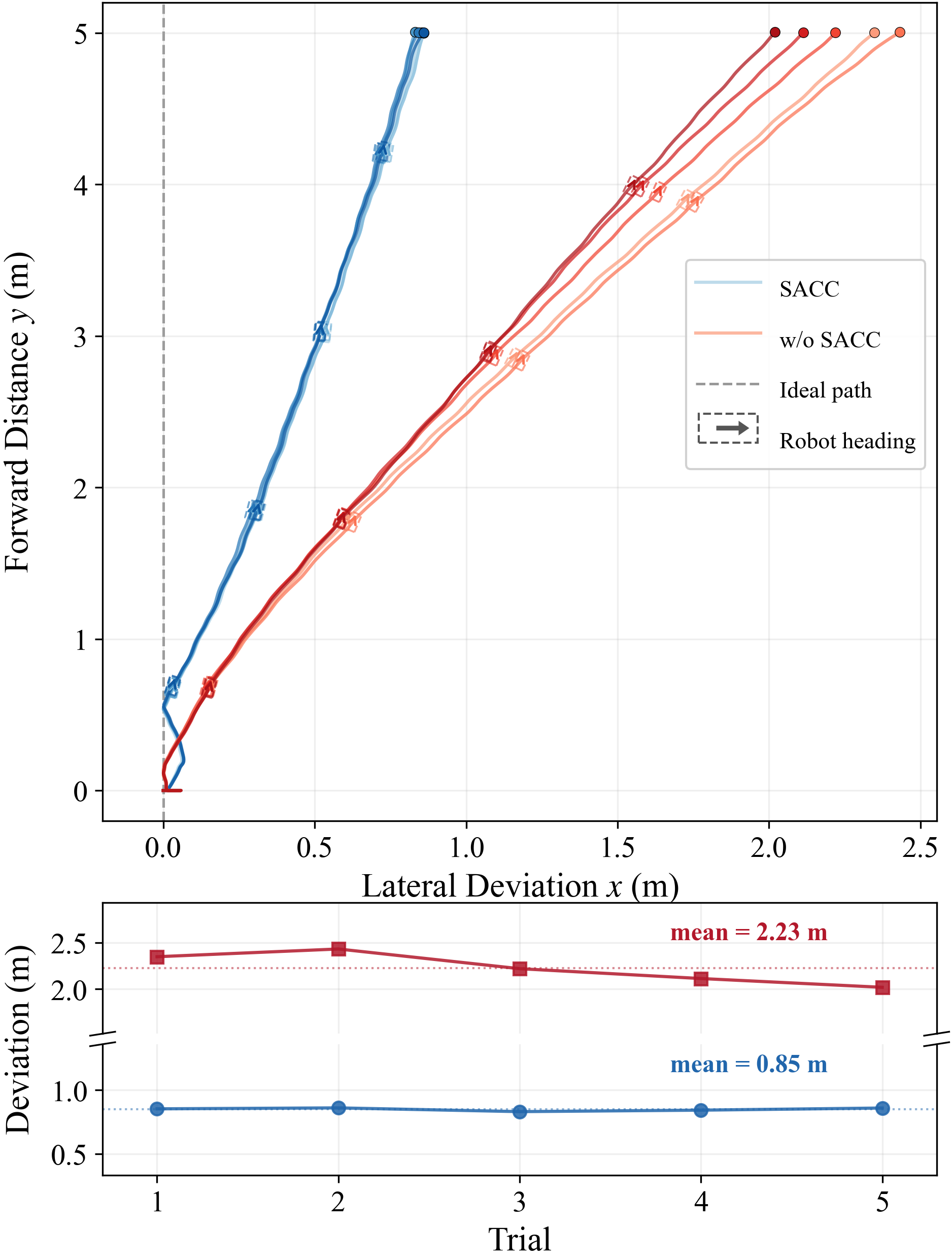}
        \caption{Straight-line tracking error along the $y$ direction (1m/s).}
        \label{fig:traj_heading_y}
    \end{subfigure}
\caption{\textbf{Trajectory tracking and robot heading comparison under straight-line velocity commands.} We compare \textbf{SACC} and \textbf{w/o SACC} under commands of \textbf{1.0~m/s} along two orthogonal directions. Dashed boxes indicate the \textbf{robot heading} at selected timestamps. SACC maintains stable heading alignment and follows near-straight trajectories with small lateral deviation in both directions, whereas \textbf{w/o SACC} accumulates heading error into curved paths.}
\label{fig:traj_heading}
\end{figure}

\textbf{Qualitative Trajectory Analysis.} We first visually assess directional generalization under 1.0~m/s straight-line commands along orthogonal axes. As illustrated in Fig.~\ref{fig:traj_heading}, the unconditioned baseline (\textbf{w/o SACC}) suffers from massive heading drifts and severe lateral deviations. It rapidly degenerates into strongly curved trajectories as the prior forcefully pulls the policy back toward the forward training manifold. In contrast, Diff-CAST maintains stable heading alignment and tracks the intended straight paths with negligible deviation.

\begin{table}[ht]
\centering
\caption{Evaluation of Omnidirectional Command Tracking}
\label{tab:tracking_performance}
\resizebox{\columnwidth}{!}{%
\begin{tabular}{l|cc|cc}
\toprule
\multirow{2}{*}{\textbf{Target Command Profile}} & \multicolumn{2}{c|}{\textbf{w/o SACC (Baseline)}} & \multicolumn{2}{c}{\textbf{Diff-CAST (Ours)}} \\ \cmidrule(l){2-5} 
 & \begin{tabular}[c]{@{}c@{}}Pos. Dev.\\ ($\downarrow$ m)\end{tabular} & \begin{tabular}[c]{@{}c@{}}Heading Drift\\ ($\downarrow$ rad)\end{tabular} & \begin{tabular}[c]{@{}c@{}}Pos. Dev.\\ ($\downarrow$ m)\end{tabular} & \begin{tabular}[c]{@{}c@{}}Heading Drift\\ ($\downarrow$ rad)\end{tabular} \\ \midrule
\textbf{Forward Walk} ($v_x = 1.0$)   & 25.03 & 2.07 & \textbf{1.08} & \textbf{0.13} \\
\textbf{High-Speed Run} ($v_x = 3.5$) & 4.55 & 0.50 & \textbf{3.29} & \textbf{0.38} \\
\textbf{Pure Lateral} ($v_y = 1.0$)   & 0.83 & 0.1 & \textbf{0.31} & \textbf{0.04} \\
\textbf{Pure Backward} ($v_x = -1.0$) & Fail (OOD) & Fail (OOD) & \textbf{0.21} & \textbf{0.16} \\
\bottomrule
\end{tabular}%
}
\end{table}

\textbf{Quantitative Omnidirectional Benchmark.} To systematically quantify this capability, \textbf{Table~\ref{tab:tracking_performance}} presents a comprehensive evaluation across a diverse velocity envelope, including high-speed running, pure lateral stepping, pure backward walking, and diagonal maneuvers. The tracking error measures the cumulative spatial deviation from the commanded path over a task-specific evaluation distance (40 m for forward, and 5 m for lateral and backward)
The unconditioned baseline heavily degrades or completely fails (e.g., severe spinning or command override) under extreme out-of-distribution (OOD) commands, such as pure backward motions. Conversely, Diff-CAST consistently achieves low position tracking deviation and near-zero heading drift across all tested conditions, demonstrating robust zero-shot multi-axis decoupling.

\subsection{Sim-to-Real Transfer and Hardware Safety Analysis}
We evaluate the effectiveness of the Safe Sim-to-Real architecture in ensuring a stable and hardware-compliant physical deployment.

\begin{figure}[h]
    \centering
    
    \begin{subfigure}{\linewidth}
        \centering
        \includegraphics[width=\linewidth]{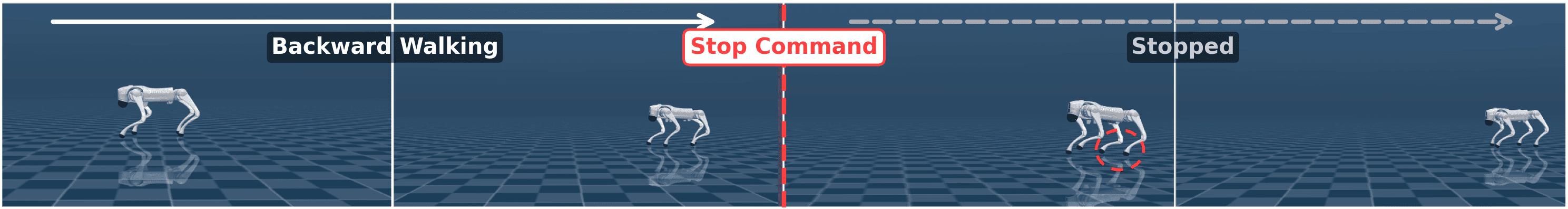}
        \caption{w/o transient-aware command curriculum}
        \label{fig:stop_a}
    \end{subfigure}
    
    \begin{subfigure}{\linewidth}
        \centering
        \includegraphics[width=\linewidth]{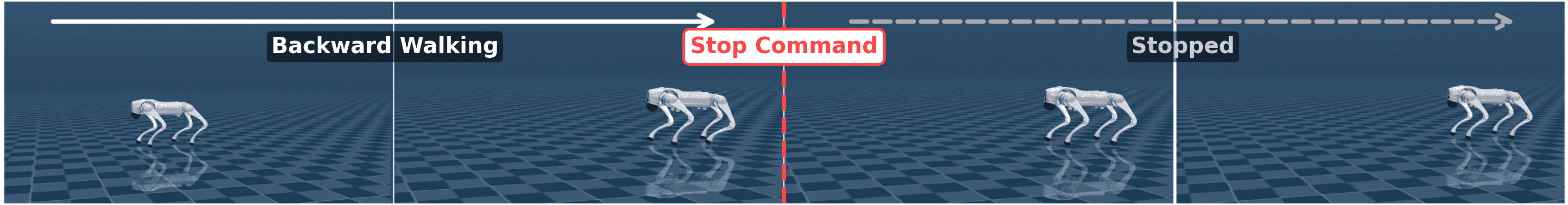}
        \caption{Diff-CAST}
        \label{fig:stop_b}
    \end{subfigure}

    \caption{\textbf{Effect of transient-aware command curriculum on emergency stop stability.} (a) The policy trained without transient-aware command curriculum exhibits instability, characterized by abnormal front leg lifting upon receiving the stop command. (b) Diff-CAST executes a stable emergency stop during backward walking. }
    \label{fig:cmd_curr}
\end{figure}

\textbf{Emergency Stop Stability. }
To assess transient stability under abrupt command changes, we perform a stress test involving an immediate stop command during backward walking. As shown in Fig.~\ref{fig:cmd_curr}(a), the baseline policy lacking the transient-aware command curriculum exhibits highly erratic and unsafe dynamics, physically manifesting as an abnormal front-leg lifting reflex. In contrast, our curriculum-guided policy enables the physical robot to learn compliant deceleration and safely transition to a stable resting posture (Fig.~\ref{fig:cmd_curr}(b)).

\begin{figure}[h] 
    \centering
    
    \begin{subfigure}{\linewidth}
        \centering
        \includegraphics[width=0.9\linewidth]{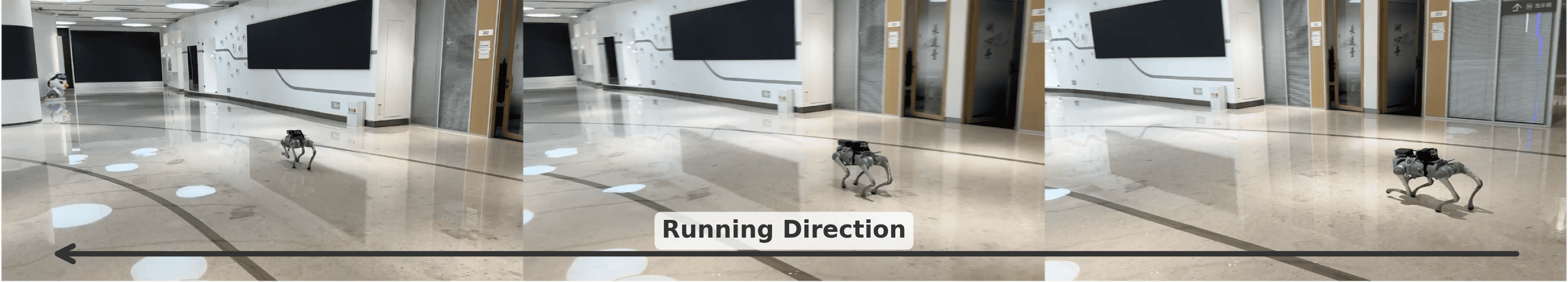} 
        \caption{Real-world sprint (3 m/s)}
        \label{fig:sub_a}
    \end{subfigure}
    
    \begin{subfigure}{0.48\linewidth}
        \centering
        \includegraphics[width=\linewidth]{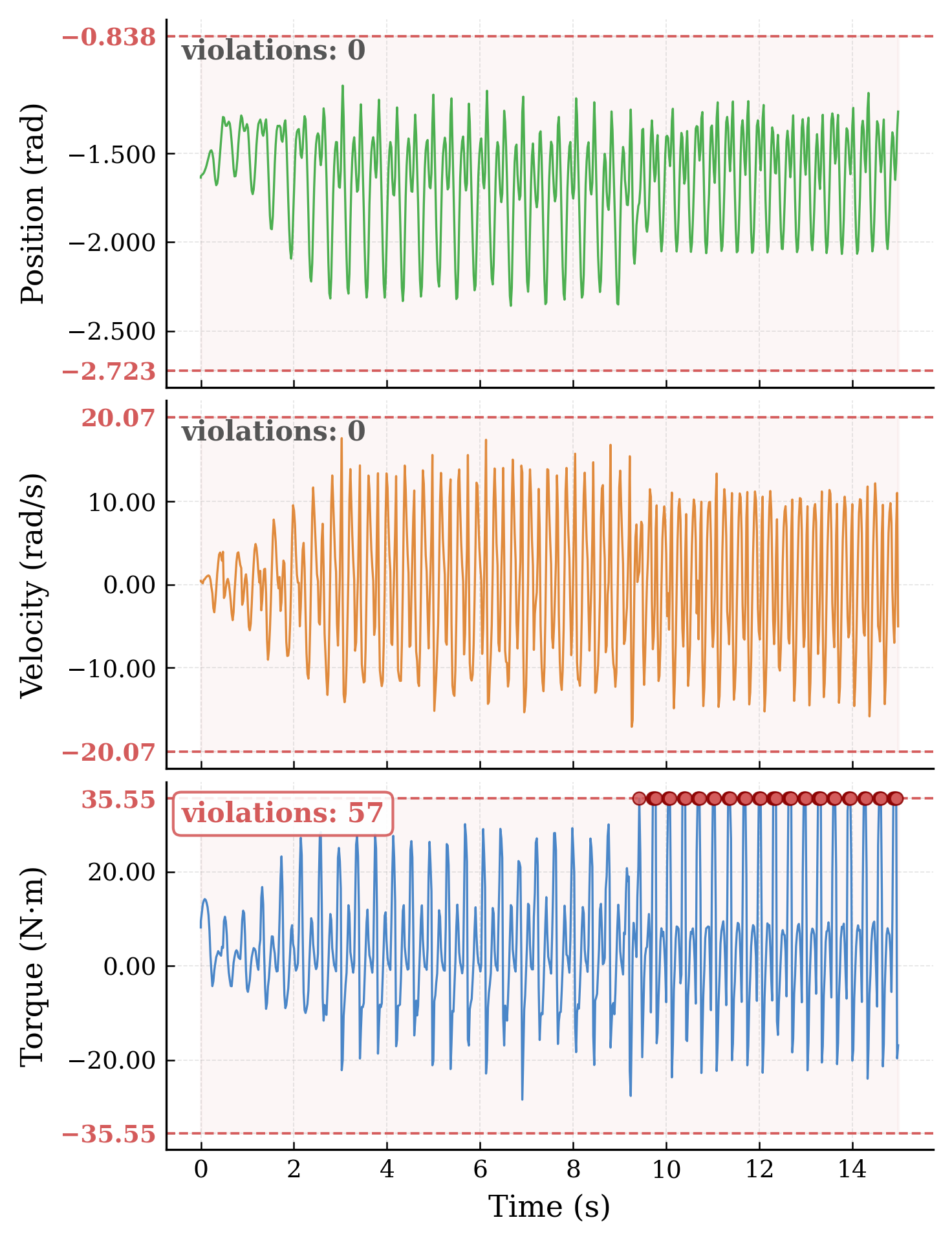}
        \caption{w/o Constrained RL}
        \label{fig:sub_b}
    \end{subfigure}
    \hfill 
    \begin{subfigure}{0.48\linewidth}
        \centering
        \includegraphics[width=\linewidth]{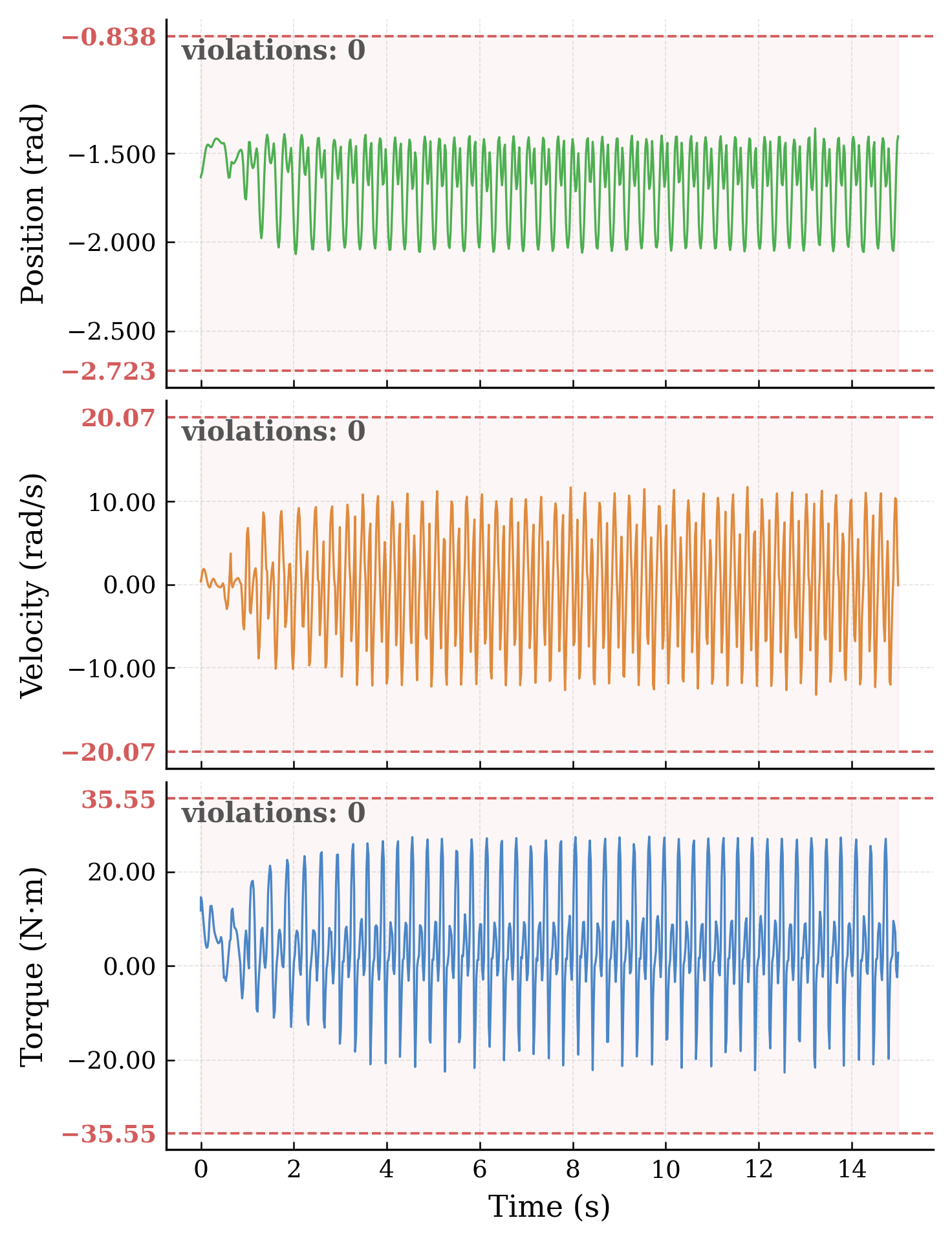}
        \caption{Diff-CAST}
        \label{fig:sub_c}
    \end{subfigure}

    \caption{\textbf{Comparison of FR calf joint dynamics during a $3\text{ m/s}$ physical run (target velocity 3 m/s maintained for 15 s).} (a) Real-world deployment scenario. (b) Without Constrained RL, the policy generates aggressive commands that repeatedly violate joint torque limits. (c) Diff-CAST strictly adheres to all physical constraints (0 violations).}
    \label{fig:sim2real}
\end{figure}

\textbf{Actuator Constraint Compliance.} We analyze the joint dynamics of the Front-Right (FR) calf during a sprint, where the robot is commanded to track a target velocity of 3 m/s and maintain it for 15 seconds (Fig.~\ref{fig:sim2real}(a)). The unconstrained baseline \textbf{(w/o Constrained RL)} generates aggressive control signals that violently and repeatedly breach the hardware safety thresholds for joint velocity and torque (\textbf{indicated by the red dashed limits and violation counters} in Fig.~\ref{fig:sim2real}(b))). This unconstrained tracking induces destructive high-frequency torque chattering (often approaching the policy control frequency), accumulating 57 safety torque limit violations. Conversely, Diff-CAST ensures all generated joint commands remain strictly within physical limits (\textbf{0 violations}, Fig.~\ref{fig:sim2real}(c)). The resulting torque profile is remarkably smooth without saturation, cleanly concentrating the control energy at the robot's macroscopic stepping frequency to guarantee highly reliable hardware-safe operation.


\begin{table}[t]
    \centering
    \caption{Quantitative Evaluation and Ablation Study}
    \label{tab:ablation}
    \begin{tabular*}{\columnwidth}{@{\extracolsep{\fill}}lccc@{}} 
        \toprule
        \textbf{Method Variant} & \textbf{Tracking Err. $\downarrow$} & \textbf{FGD $\downarrow$} & \textbf{Safety Viol. $\downarrow$} \\ 
        \midrule
        Vanilla AMP & 26.85 & 4173.66 & Fail \\
        \midrule
        \textbf{Diff-CAST (Ours)} & \textbf{1.08} & 489.13 & \textbf{0} \\
        $\quad\rightarrow$ \textit{w/o SACC} & 25.03 & \textbf{348.55} & \textbf{0} \\ 
        $\quad\rightarrow$ \textit{w/o Safe Sim2Real} & 12.58 & 1845.20 & 57 \\
        \bottomrule
    \end{tabular*}
    
    \vspace{3pt} 
    \raggedright \small \textit{Note:} $\downarrow$ lower is better. Safety violations are measured during a 15 s sprint at 3 m/s.
\end{table}

\subsection{Summary of Quantitative Ablations}
\label{subsec:quantitative_ablations}
To isolate the contribution of each proposed component, we consolidate the quantitative ablation results across all metrics evaluated in Table~\ref{tab:ablation}. In summary, the full Diff-CAST framework is essential to achieve an optimal balance between kinematic fidelity and versatile control. Specifically, the diffusion prior drastically improves motion realism, reducing the FGD by nearly an order of magnitude compared to the vanilla baseline. Although the w/o SACC variant yields a numerically lower FGD, it does so by passively overfitting to the forward-biased expert data—completely ignoring out-of-distribution user commands, which inflates the tracking error to a severe 25.03. SACC successfully resolves this objective conflict, guaranteeing precise omnidirectional tracking (lowest error) while retaining highly realistic gaits. Finally, the Safe Sim-to-Real architecture remains strictly indispensable for hardware-compliant deployment (zero safety limit violations).

\section{CONCLUSIONS}
Diff-CAST effectively resolves mode collapse and objective conflicts in data-driven biomimetic locomotion by replacing conventional binary discriminators with a bounded, action-agnostic diffusion prior. Our framework unifies heterogeneous motor skills into a single versatile policy. Symmetric Augmented Command Conditioning (SACC) mathematically resolves OOD task-style mismatch, eradicating heading drifts for precise omnidirectional tracking. Finally, our Constrained RL architecture strictly prevents torque violations, enabling fluid, hardware-safe multi-modal transitions on a physical quadruped. Future work will explore integrating exteroceptive vision to safely encapsulate highly explosive, underactuated aerial maneuvers.



\bibliographystyle{IEEEtranS}
\bibliography{IEEEfull}

\end{document}